**Title:**

Machine learning for prediction of dose-volume histograms of organs-at-risk in prostate cancer from simple structure volume parameters.

**Authors:**

Saheli Saha
Department of Radiation Oncology, Tata Medical Center, Kolkata, West Bengal, India

Debasmita Banerjee
Department of Mathematics, Indian Institution of Technology, Patna, India

Rishi Ram
Department of Mechanical Engineering, Indian Institute of Technology (BHU), Varanasi, India

Gowtham Reddy
School of Medical Science and Technology, Indian Institution of Technology, Kharagpur, West Bengal, India

Debashree Guha
School of Medical Science and Technology, Indian Institution of Technology, Kharagpur, West Bengal, India

Arnab Sarkar
Department of Mechanical Engineering, Indian Institution of Technology (BHU) Varanasi, India

Bapi Dutta
The Logistics Institute - Asia Pacific, National University of Singapore, Singapore

Moses ArunSingh S
Department of Radiation Oncology, Tata Medical Center, Kolkata, West Bengal, India

Suman Chakraborty
Department of Mechanical Engineering, Indian Institute of Technology Kharagpur, West Bengal, India

Indranil Mallick
Department of Radiation Oncology, Tata Medical Center, Kolkata, West Bengal, India

**Corresponding Author**: Indranil Mallick
**Mailing Address**: 4, MAR(E-W), DH Block(Newtown), Action Area I, Newtown, Kolkata, West Bengal 700160

**Email Address:** imallick@gmail.com

**Abstract:**
Dose prediction is an area of ongoing research that facilitates radiotherapy planning. Most commercial models utilise imaging data and intense computing resources. This study aimed to predict the dose-volume of rectum and bladder from volumes of target, at-risk structure organs and their overlap regions using machine learning. Dose-volume information of 94 patients with prostate cancer planned for 6000cGy in 20 fractions was exported from the treatment planning system as text files and mined to create a training dataset. Several statistical modelling, machine learning methods, and a new fuzzy rule-based prediction (FRBP) model were explored and validated on an independent dataset of 39 patients. The median absolute error was 2.0%-3.7% for bladder and 1.7-2.4% for rectum in the 4000-6420cGy range. For 5300cGy, 5600cGy and 6000cGy, the median difference was less than 2.5% for rectum and 3.8% for bladder. The FRBP model produced errors of 1.2%, 1.3%, 0.9% and 1.6%, 1.2%, 0.1% for the rectum and bladder respectively at these dose levels. These findings indicate feasibility of obtaining accurate predictions of the clinically important dose-volume parameters for rectum and bladder using just the volumes of these structures.

**Introduction:**
Radiation therapy (RT) using intensity-modulated radiation therapy (IMRT) is a commonly used and cost-effective approach to treating localised and low-volume oligometastatic prostate cancer[1]. IMRT is required in prostate cancer RT to deliver a high dose of RT to the prostatic tumour while minimising the dose to the adjacent organs at risk (OARs), namely the rectum and bladder.

A method of upfront prediction of the doses that will be achieved by this planning process (to the target volumes and OARs) would be useful on many counts. It would enable the planner to take steps in advance to improve plans if necessary and would help the clinician discuss expectations with the planner and the patient.

Prior studies have evaluated models for predicting dose-volume histograms (DVH) and dose distribution using anatomical information in the planning CT scan and a library of prior plans. Commercially available systems have used the term knowledge-based planning (KBP) for some of these approaches. Several knowledge-based prediction models have been validated[2–4]. By and large, these solutions are dependent on volumetric images and anatomical segmentations in DICOM or equivalent format as input parameters for model generation and prediction. They use deep-learning approaches, which are resource-hungry. Some solutions, like RapidPlan$^{(TM)}$, are vendor-specific.

This study aimed to simplify this process and test if a DVH prediction can be achieved for the urinary bladder and rectum, using just the volumes (in cc) of the target structures, the OARs, and their overlap regions as input parameters.

As a proof of concept, several standard machine learning regression approaches were used, and a fuzzy rule-based prediction model integrated with the fuzzy decision tree (FDT) as a comparable alternative to the numerical learning techniques was also trained and tested[5–7]. The goal was to develop, compare and validate DVH prediction models using machine learning with a resource-sparing, scalable, and interpretable solution.

**Materials and Methods:**
**Datasets for model training and validation:**

A retrospective dataset of clinically approved plans of patients of localised prostate cancer treated between 2015 and 2017 with curative-intent moderately hypofractionated radiotherapy to a dose of 6000cGy in 20 fractions to the planning target volume (PTV) was selected for training. All patients had been first delineated on the Varian (™) Eclipse system and were subsequently planned on either the Tomotherapy (™) or Eclipse treatment planning systems. Clinical approval for treatment had been made based on published institutional dose constraints derived from the QUANTEC guidelines and published hypofractionation studies[8–11].

An initial dataset of 94 approved plans was selected for training and tuning the models. Using a random 7:3 split, a 'training' dataset of 65 plans, and a 'test' dataset of 29 plans, the latter being used for hyperparameter tuning. In addition, an independent 'validation' set of 39 patients was also selected from other patients treated during the same period for testing the prediction accuracy. All of these cases had been delineated using the standard contouring guidelines and planned and treated previously using the same dose constraints as those in the training and test set. The trained models were then tested on the validation dataset with the model binaries and pre-defined regression metrics, as mentioned below.

The study was approved by the institutional review board (2019/TMC/144/IRB9).

**Data preparation:**

The process flow is documented in Figure 1 and designed to accommodate more than one treatment planning system. To build the training and testing library, DVH data were exported from the treatment planning systems (TPS). The cumulative dose-volume (percentage/cc) details with a dose bin of 10cGy were exported as text files from the Varian Eclipse treatment planning system and as comma-separated value (CSV) files from the Accuracy Tomotherapy system. Any patient-identifying information was removed.

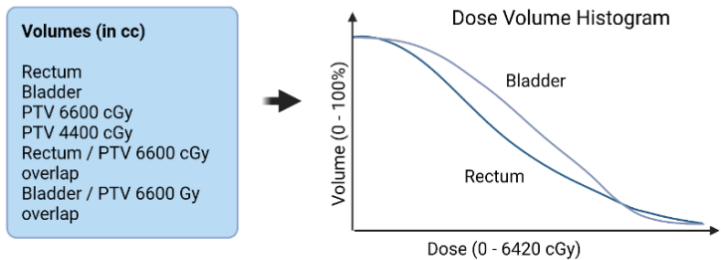

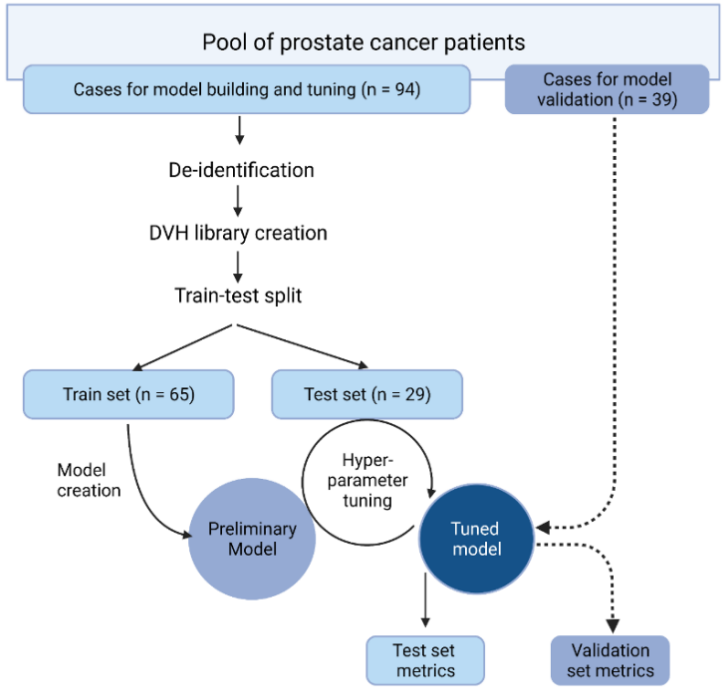

Fig 1: Goals and Process Flow of Modelling. The goals were to predict the DVH from the volumes of the 6 structures. The initial de-identified data set was split into a training and a test set. The initial models were tuned using hyperparameter tuning to form a final model that was validated on an independent set

Using rule-based text-mining, the dose-volume information of the following structures was extracted - the PTV receiving the prescription dose of 6000cGy, the PTV receiving the elective pelvic dose of 4400cGy, the urinary bladder, the anorectum, and the overlap volume of the bladder and rectum with the PTV receiving 6000cGy. This data was collated into organ-specific dose-volume libraries. The structure volumes of the six structures were extracted into a feature library for training and have been used as the input features.

A formal sample size calculation was not performed. As a rule of thumb in machine learning tasks, at least 10 events/cases are required per predicting parameter [12]. Hence, with six input features, a sample size of 60 or higher was considered adequate for the training set. For a linear regression task with 6 predictors with an alpha error of 0.05 and a power of 0.8, the effect size that can be determined with 65 patients used in the training set is 0.23 (G*power v 3.1.9.6).

**Model training:**

Appreciating that different regression methods have different advantages and drawbacks[13] and that this was the first implementation in DVH prediction, this study aimed to explore different regression models for accuracy. Ten different models (including linear models, tree-based ensemble models, boosting methods, support vector regression (SVR) and multilayer perceptron (MLP) as listed in Table 1) were employed. Using the 6 features, these models were trained for predicting percentage volumes for 642 doses at 10cGy intervals from 0 to 6420cGy. Model hyperparameters were tuned using grid-search cross-validation.

| **Input parameters** | PTV60Gy volume, PTV44Gy volume, Rectum volume, Bladder volume, Overlap volume of Rectum and PTV60Gy (as a fraction of the Rectal volume), Overlap volume of Bladder and PTV60Gy (as a fraction of the Bladder volume) |
|---|---|
| **Models used** | Linear Regression(LR), Decision Tree(DT), Random Forest(RF), Elastic Net(EN), Gradient Boosting Regressor(GBR), Extreme Gradient Boosting(XGBR), Support Vector Regression(SVR), MultiLayer Perceptron(MLP), Fuzzy rule-based prediction(FRBP) model, Monte Carlo method(MCM), Ensemble of 6 and 3 best methods |

Table 1: Parameters and models used for training

In addition to the above-mentioned algorithms, a fuzzy rule-based prediction model (FRBP) integrated with the fuzzy decision tree was generated and tested. This is the first implementation of this approach in DVH prediction. In comparison to other machine learning methods, the FRBP method has the potential to avoid overfitting and in achieving a reasonable level of accuracy[6]. The FRBP model essentially creates fuzzy partitions based on the cluster centres of data points and merged the partitions (instead of modifying the data points of the clusters) by aggregating the membership functions of the fuzzy partitions so that the partitions generated by the method are not independent of each other. The optimal fuzzy partitions are selected with small standard deviations, i.e., more homogeneity, with the view of reducing the variance within the data points of each partition. Finally, linguistic labels to the fuzzy sets associated with the optimal fuzzy partition are assigned to create an interpretable fuzzy rule base in order to predict the result by employing the fuzzy decision tree approach. A complete and accurate fuzzy rule base improves the system's interpretability and accuracy[14]. Further details of this method are provided in the Appendix A. In addition, an ensemble by taking an average of predictions of the outperforming models was also created to test if the accuracy and reliability could be improved over individual models.

**Model assessment:**

The median absolute error (MAE) was used as the main metric to measure the model prediction performance because it is not sensitive to outliers. It can be quantified by the median of the absolute differences between the actual and predicted volumes as given below:

MAE $(V_t, V_p)$ = median $(|V_{1t} - V_{1p}|, |V_{2t} - V_{2p}|, …, |V_{nt} - V_{np}|)$

where, $V_t = \{V_{1t}, V_{2t}, …, V_{nt}\}$ was the sample consisting of true volumes and $V_p = \{V_{1p}, V_{2p}, …, V_{np}\}$ was the corresponding sample consisting of predicted volumes at n dose points respectively. The overall deviation of predicted DVHs from the clinically planned DVHs was estimated by the average MAE over entire patients.

The values of MAE were calculated for a) clinically important individual point doses used as a standard in our practice (5300 cGy, 5600cGy, 6000cGy); b) over the entire dose range (averaged) as well as c) over prespecified dose bins divided into low (0-1990cGy), intermediate (2000-3990cGy) and high dose (4000-6420cGy) levels (averaged) on 'test' datasets. Based on regression performance, we selected the best models from tested algorithms and saved the model binaries for testing on the validation dataset.
The Kruskal Wallis test was performed to calculate the variance between actual and predicted doses across different models, with *p-value <0.05* considered significant[15]. The statistical summary is provided in Appendix B.

The confidence interval (95% degree of confidence) was estimated through the Weibull model (Appendix C) using the training data set.

To improve the overall performance and stability of the prediction results, attempts were made to combine the decisions of the best-performing methods in ensemble models. The ensemble learning method combines the strengths of the individual models with a view to creating a better model for prediction. Two different ensemble approaches were experimented upon, one by combining all 6 models in an ensemble and another by considering only the 3 best learning models based on their MAE and variance. The final prediction of the ensemble model was done by taking an average of predictions from all the models.

**Tools:**

The text data extraction and regression modelling, and performance evaluation were done using Python 3.8 and its NumPy, pandas, matplotlib, and sci-kit-learn libraries. The fuzzy rule-based learning was developed on MATLAB vR2015a and Python 2.7.16.

**Reporting:**

The TRIPOD checklist was used when writing this report.

**Results:**

The calculated MAE for both urinary bladder and rectum in the test and validation sets and in the different dose ranges with different models have been summarised in Table 2. The average MAE was less than 5.3% for bladder and less than 5% for rectum in the validation set for the entire dose range. The average MAE was lower in the clinically more important dose range of 4000-6420 cGy, ranging from 2.0% to 3.7% for bladder and 1.7% to 2.4% for rectum for most of the models. However, there was greater MAE in the dose range of 2000-3990 cGy.

| Median Absolute Error in percentage volume receiving given doses | | | | | | | | |
|---|---|---|---|---|---|---|---|---|
| BLADDER | | | | | | | | |
| METHOD | DATASET | 0-6420 cGy | 0-1990 cGy | 2000-3990 cGy | 4000-6420 cGy | 5300 cGy | 5600 cGy | 6000 cGy |
| | | (avg over range) | (avg over range) | (avg over range) | (avg over range) | | | |
| **LR** | Test | 5.1 | 4.2 | 10.3 | 1.7 | 1.5 | 1.1 | 0.9 |
| | Validation | 4.6 | 3.3 | 9.8 | 1.4 | 1.3 | 0.8 | 0.6 |
| **RF** | Test | 4.3 | 1.8 | 9.8 | 2.0 | 1.8 | 1.4 | 0.9 |
| | Validation | 3.8 | 1.2 | 8.6 | 1.5 | 1.4 | 1.1 | 0.8 |
| **DT** | Test | 4.2 | 0.7 | 6.4 | 1.5 | 2.8 | 2.0 | 1.2 |
| | Validation | 3.7 | 0.7 | 9.6 | 1.9 | 1.4 | 1.3 | 0.6 |

| METHOD | DATASET | | | | | | |
|---|---|---|---|---|---|---|---|
| **GBR** | Test | 4.5 | 2.0 | 9.9 | 2.3 | 2.0 | 1.5 | 0.9 |
| | Validation | 3.9 | 1.9 | 8.2 | 1.9 | 1.6 | 1.3 | 0.8 |
| **XGB** | Test | 4.3 | 1.5 | 9.6 | 2.2 | 2.2 | 1.2 | 0.7 |
| | Validation | 3.5 | 1.1 | 8.3 | 1.5 | 1.4 | 1.1 | 0.8 |
| **FRBP** | Test | 4.7 | 2.0 | 10.5 | 2.1 | 1.6 | 1.3 | 1.1 |
| | Validation | 3.8 | 0.4 | 9.4 | 2.0 | 1.7 | 1.2 | 0.7 |
| **EN** | Test | 5.4 | 4.2 | 11.0 | 1.7 | 1.4 | 1.0 | 1.1 |
| | Validation | 4.4 | 2.7 | 9.7 | 1.4 | 1.4 | 1.0 | 0.9 |
| **SVR** | Test | 4.0 | 1.9 | 9.0 | 1.7 | 1.3 | 1.0 | 0.6 |
| | Validation | 5.3 | 1.5 | 12.8 | 2.3 | 2.3 | 1.5 | 1.1 |
| **MLP** | Test | 5.2 | 4.2 | 10.3 | 1.7 | 1.5 | 1.1 | 0.9 |
| | Validation | 4.6 | 3.3 | 9.8 | 1.4 | 1.3 | 0.8 | 0.6 |
| **MCM** | Test | 10.6 | 15.2 | 20.9 | 5.3 | 5.0 | 3.9 | 1.7 |
| | Validation | 12.5 | 15.6 | 24.5 | 6.4 | 6.3 | 5.0 | 2.6 |
| **RECTUM** | | | | | | | |
| METHOD | DATASET | 0-6420 cGy | 0-1990 cGy | 2000-3990 cGy | 4000-6420 cGy | 5300 cGy | 5600 cGy | 6000 cGy |
| | | (avg over range) | (avg over range) | (avg over range) | (avg over range) | | | |

| | | | | | | | |
|---|---|---|---|---|---|---|---|
| **LR** | Test | 4.1 | 2.4 | 8.3 | 2.0 | 1.5 | 1.6 | 1.1 |
| | Validation | 4.2 | 3.4 | 8.2 | 1.7 | 1.1 | 1.3 | 1.0 |
| **RF** | Test | 5.1 | 3.0 | 10.5 | 2.1 | 0.7 | 1.2 | 1.1 |
| | Validation | 3.9 | 3.2 | 7.2 | 1.6 | 1.0 | 1.2 | 0.9 |
| **DT** | Test | 5.8 | 3.4 | 12.2 | 2.2 | 2.0 | 1.1 | 1.0 |
| | Validation | 4.3 | 3.7 | 8.0 | 2.0 | 0.9 | 1.8 | 1.0 |
| **GBR** | Test | 5.7 | 3.6 | 11.6 | 2.6 | 1.8 | 1.6 | 1.3 |
| | Validation | 4.5 | 3.9 | 8.3 | 1.8 | 1.4 | 1.5 | 1.1 |
| **XGB** | Test | 5.3 | 3.7 | 10.3 | 2.5 | 1.1 | 1.2 | 1.2 |
| | Validation | 4.7 | 3.9 | 9.0 | 1.8 | 1.3 | 2.3 | 0.9 |
| **FRBP** | Test | 5.3 | 2.7 | 11.2 | 2.6 | 1.6 | 1.7 | 1.1 |
| | Validation | 4.0 | 3.2 | 7.2 | 2.0 | 1.2 | 1.3 | 0.9 |
| **EN** | Test | 4.4 | 2.8 | 9.2 | 1.7 | 1.1 | 1.6 | 1.2 |
| | Validation | 4.2 | 3.5 | 7.8 | 1.7 | 1.7 | 1.7 | 1.0 |
| **SVR** | Test | 4.1 | 1.9 | 8.3 | 2.5 | 1.9 | 1.3 | 1.0 |
| | Validation | 4.5 | 4.0 | 8.7 | 1.6 | 1.6 | 1.5 | 0.7 |
| **MLP** | Test | 4.1 | 2.4 | 8.3 | 2.0 | 1.5 | 1.6 | 1.1 |
| | Validation | 4.2 | 3.4 | 8.2 | 1.7 | 1.1 | 1.3 | 1.0 |
| **MCM** | Test | 9.7 | 12.3 | 21.5 | 4.9 | 4.6 | 3.3 | 2.9 |

|  | Validation | 7.1 | 8.5 | 16.1 | 4.6 | 5.1 | 3.9 | 1.5 |

Table 2: Error metrics for bladder and rectal dose prescription, using dose ranges and specific doses. Median Absolute Errors (MAE) for test and validation sets for bladder and rectum at full dose range, low (0-1990 cGy), intermediate (2000-3990 cGy), high dose range (4000-6420 cGy) and specific dose points.

For both urinary bladder and rectum LR, RF, EN, FDT, SVR, MLP methods outperformed MCM, DT, GBR, and XGB in terms of accuracy without overfitting. Overfitting was determined by the difference in accuracy between test and validation sets.

Using the Kruskall Wallis test, we did not find a significant difference in the average of actual vs. predicted volumes over the entire dose range, and in individual doses (Tables 2 and 3, Appendix).

Using the confidence intervals generated from the Weibull model, it was confirmed that the predicted DVH from the tested models lay between confidence intervals for both the bladder and rectum (Figures 4a and 4b, Appendix).

The variances in MAE were compared in the validation cases for bladder and rectum at the high dose range as well as the full dose range for the above-mentioned better-performing methods (shown in Figure 2). For bladder, LR, MLP, EN, and FRBP showed the least variation between actual and predicted doses, while for rectum, RF, MLP, and FRBP showed the least variance.

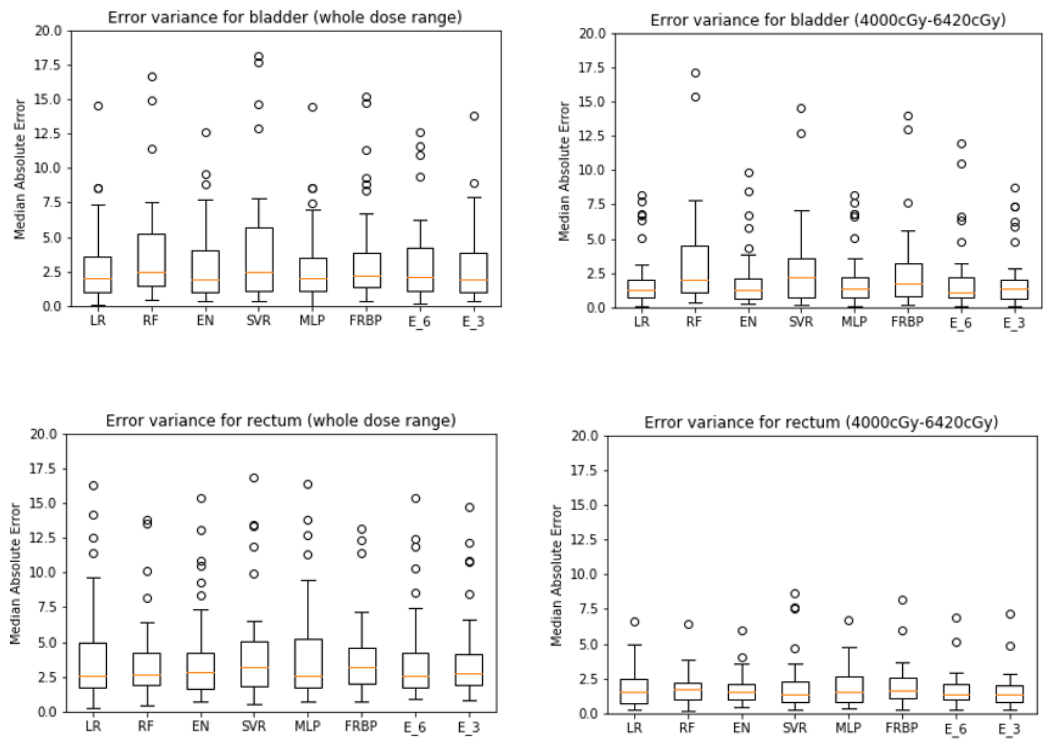

Fig 2: Error Variance for bladder and rectum across full and high dose ranges using 6 different methods with ensemble of 6 and 3 best models. Box and Whisker plots comparing the median absolute errors (MAE) of the better performing models and their ensembles for entire and high dose ranges of bladder and rectum.

Two different ensemble methods, averaging the 3 best and the 6 best models, were used. The average MAE for both the ensemble methods was less than that for individual models when the entire dose range was considered but did not result in improvement in the higher dose range. The error variances were similar (as depicted in Figure 2). There was no clear winner among the different models, as several methods and the ensembles had accurate results in the clinically important higher dose range.

Figures 3a and 3b depict the cumulative DVH as predicted by the different algorithms with the actual DVH of some of the representative cases from the external validation set, demonstrating the range of predicted vs. actual DVH.

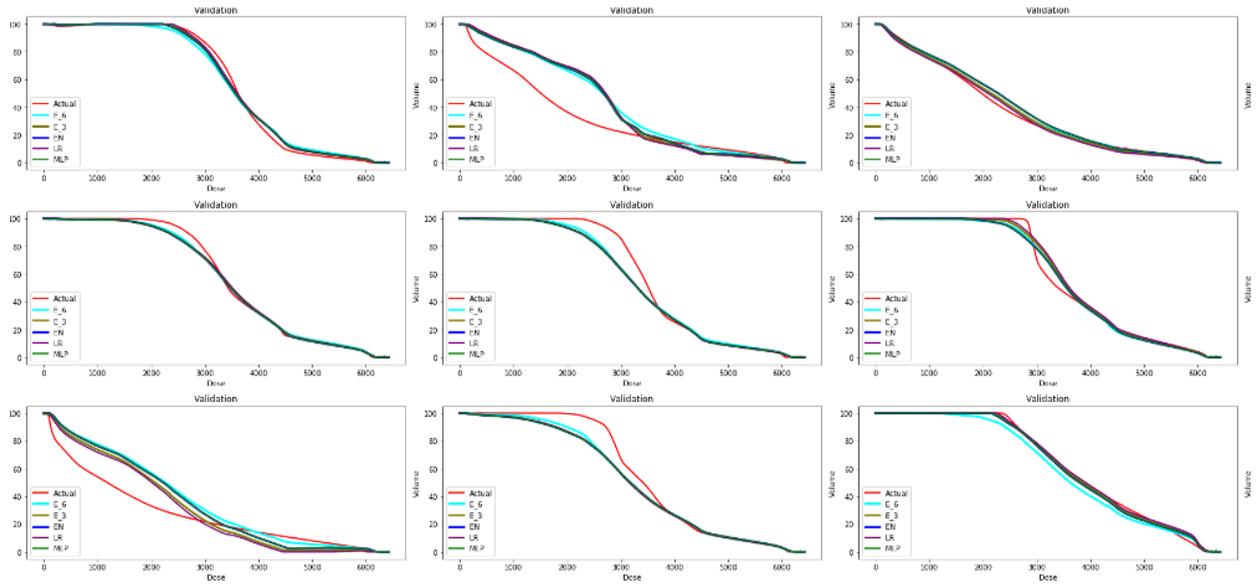

Fig 3a: Examples of cumulative DVH predicted vs actual of the bladder in 9 patients in the external validation cohort.

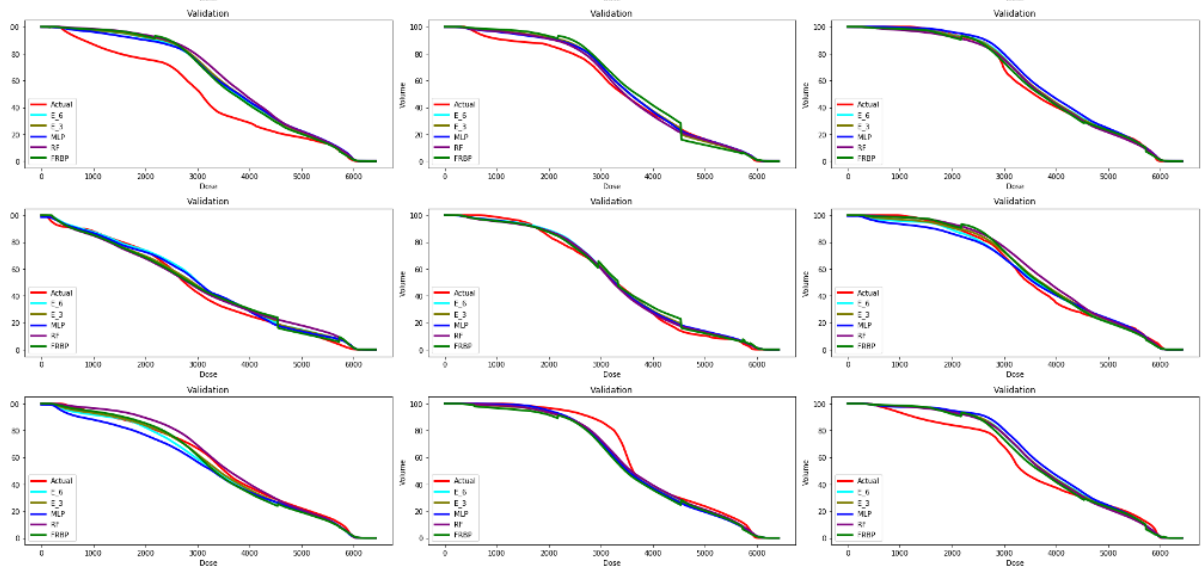

Fig 3b: Examples of cumulative DVH predicted vs actual of the rectum in 9 patients in the external validation cohort.

**Discussion:**

To the best of our knowledge, this is the first attempt to predict DVH for OARs in prostate cancer IMRT using machine learning with only the measure of the volumes of the PTV, organs at risk and their overlap volumes, without the direct use of images and segmentation masks.
Individual models for 9 different supervised regression algorithms were developed, and a decision tree was based on a fuzzy rule base. Different models with varying computational complexity were selected with the aim of assessing the performance of the relatively simpler models with the other methods. Based on a selection of the best models, we also created ensembles of six best and three best models.

It was possible to achieve accurate DVH predictions with the MAE within 2-3% of the volumes in the higher, more clinically relevant dose range of 4000-6420cGy. For the individual dose points at the 5300 cGy, 5600 cGy and 6000 cGy levels, the MAE was within 2%. Comparing models, it was found that the RF method showed superior prediction performance compared with a single decision tree-based model. Since RF trains the model by generating several decision trees in parallel with bootstrapping followed by aggregation, thus it can generalize over the data in a better way. Both the boosting algorithms, GBR and XGB performed poorly in bladder and rectum predictions due to overfitting, as there is a difference in prediction performance between training and test/validation datasets. The performance of the linear regression-based model was similar to the RF, EN, MLP, and FRBP. The models were robust in as much as their error estimates performed well on the validation dataset.

No system could outperform another, and several models provided accurate results Creating ensemble models did not have a clear impact on the accuracy of the prediction. This suggests that with a relatively simple set of predictive factors, changing the model has a limited impact on the predicted output, especially in the high dose range.

The fuzzy rule-based prediction with the fuzzy decision trees performed well in the entire dose range, performing comparably on the evaluation metrics used, with lower error variance in the whole dose range. In this method, articulation of the numerical data by means of linguistic labels/ fuzzy sets has the potential to make the data interpretability easier. In addition, it has the ability to generalize the output from the training data to model unseen situations. To our best knowledge, this is the first paper in this context comparing multiple methods. Also, this is the first application of fuzzy rule base in predictive modelling of DVH in radiotherapy. This method can therefore be further explored in this domain.

The expanding literature on machine learning has explored different techniques and metrics[16,17]. The majority of it comprises supervised methods using training on prior optimal plans. Some methods have used volumes delineated on the planning CT scans to estimate the overlap volume histogram (OVH) by plotting the percentage of OAR volume overlapping with an isotropically expanded or contracted target and the Euclidean distance of the OARs from the target or by calculating the fractional volume of the OAR within a certain distance from the PTV surface[18,19]. These studies have demonstrated that predicted plans are comparable to plans generated by expert planners. In some instances, the coverage and homogeneity achieved by the assisted plans were better in some parameters than those by a planner[18]. The DVH endpoints parameters evaluated were as varied as the metrics used. Table 3 summarises and compares the endpoints used in the literature and their analyses with that of the present article.

| Article | Input | Method Type | Output | Evaluation metric | Comments |
|---|---|---|---|---|---|
| Schubert et al. 2017 [4] | Clinical plan data | Rapidplan(TM) | Definition of constraints and objectives | Differences of the benchmark objectives between the clinical plan and RapidPlan based plan | Satisfactory optimization using the model |
| Zhang et al. 2018 [13] | Features derived from clinical plan data | Ensemble of 4 regression models using model stacking method | Optimal achievable DVH | RMSE | Models built on limited training data, robust to dosimetric and anatomical outliers. |
| Cagni et al. 2017 [20] | Helical Tomotherapy plans | Rapidplan(TM) | Generation of new RapidArc and Helical Tomotherapy plan | Dosimetric and radiobiological indexes | Inter-technique and inter-system application |
| Ma et al. 2019 [21] | PTV-only plans | Support vector regression (SVR) approach | Predicts DVH of clinical plan | Percentage of plans within a 10% error bound of selected doses. | Requires an intermediate step of generating a PTV-only plan. |

| Study | Input data | Method | Output | Validation metric | Findings |
|---|---|---|---|---|---|
| Zhu et al. 2011 [22] | Organ volumes and distance-to-target histogram (DTH) obtained by principal component analysis | Support vector regression (SVR) approach Multivariate stepwise regression Model/regression tree | DVH | Percentage within prediction bounds | Anatomical information quantified with less number of variables |
| Appenzoller et al. 2012 [23] | Anatomical and dosimetric data | Mathematical model based on expected dose to the minimum distance from a voxel to the PTV surface. | Predicted cumulative DVH | Average sum of residuals | Suboptimal models removed from the training dataset resulted in refined model, rather than average model |
| Fogliata et al. 2014 [24] | Geometry Based Expected Dose (GED) histograms, OAR volume, overlap volume with targets, out-of-field volume, target volume | Rapidplan(TM) | DVH | Average differences within one standard deviation for pre-specified volumes | Systematic improvement over the clinical plan |
| Powis et al. 2017 [25] | ROI volumes, PTV-OAR overlap volumes, OAR mean doses (including average rectum dose), DVH data for PTV and OAR structures and plan quality/complexity metrics- | Mathematical model (Relative model excess) | Average dose to the rectum normalised to the prescription dose plotted against fraction of the total rectal volume overlapping with 59.2Gy PTV | Relative model excess | Possible decrease in bladder and rectum doses compared to previous plans |
| | conformity index and total plan MU | | | | |
| Present Study | DVH | Linear Regression Decision Tree Random Forest Gradient Boost Extreme Gradient Boost Elastic Net Multilayer Perceptron Support Vector Regression Fuzzy rule-based prediction model Ensemble | Prediction of DVH | Median absolute error for specific dose levels and dose ranges | Less labor and resource intensive |

Table 3: Summary of studies on DVH prediction literature and comparison of the endpoints and methods with the present study

While the emerging research on dosimetric prediction has seen increasing use of images, deep-learning-based computation, the present study aimed to find a simple, reproducible technique that could be reproduced without requiring time and resource-hungry methods. The models were based on simple volume inputs, which can be easily readout from the TPS after segmentation. Only six values corresponding to the anatomical volume of the PTVs, rectum, bladder, and overlap volumes of these structures with the PTV60 were used as parameters. In contrast to the increasing use of deep learning models developed using DICOM images and DICOM-RT datasets, the method described in the present study was considerably simpler and faster and has a negligible cost.

Using the model binaries, the DVH of a new patient can be predicted within a few seconds of completing segmentation and determining the structure volumes. This can be achieved using standard computing infrastructure without the requirement for graphics processing units (GPU) and within the available computing power of any radiation oncology department in the world. This makes the presented approach more distributable and scalable in low-resource settings. A created prototype web-based dashboard for identifying DVHs of similar cases based on a library of plans has been created[26]. A similar dashboard is being built to demonstrate machine-learning predicted plans which can be accessed from any computer connected to the internet.

The dose-volume information available with the model predictions can help inform action from the perspective of the dosimetrist and the clinician. Predicted dose volumes outside recommended constraints could lead to changes in planning parameters *a priori*, rather than in reaction to poor first plan iterations, or prompt a discussion between the radiation oncologist and patient regarding expected toxicities. Alternatively, if the dose volumes predicted are optimal, standard plan templates could be used to save time for dosimetrists.

The tested approach also has certain limitations. The DVH prediction model does not provide either voxel-based doses or hardware parameters and is not meant to be a replacement for the treatment planning process. The training data is built on plans done within a single department using similar constraints and planning parameters. In such a situation, the models are specific to the department and may have limited validity in an external dataset from a different institution. However, in principle, the methodology can be followed by other individual departments to build local models very easily that are likely to provide equivalent accuracy within the local setting. The prediction accuracy in the middle range of doses 2000-3990cGy is lower than in the lower or higher dose range with this dataset. This could be due to greater variance in doses at these levels. As our departmental dose constraints are based on extrapolation of the QUANTEC constraints to a moderate hypofractionation schedule, we have not used specific constraints below 4700cGy in a 20-fraction schedule. This has resulted in more heterogeneity at lower dose volumes. It is likely that with routine use of constraints at lower dose levels, the variability will be lower.

**Conclusion:**

This study demonstrates that a simple machine learning approach to the prediction of DVH can yield accurate results and has the potential for wide application in the setting of prostate cancer IMRT. A fuzzy rule-based partitioning with decision-tree approach can be a useful and interpretable addition to the repertoire of conventional modelling algorithms.

**Acknowledgement:**

The work is supported by the grant of SERB, Government of India (Ref.No:SPG/2022/000045**).**

**References:**

[1] Noble SM, Garfield K, Athene Lane J, Metcalfe C, Davis M, Walsh EI, et al. The ProtecT randomised trial cost-effectiveness analysis comparing active monitoring, surgery, or radiotherapy for prostate cancer. Br J Cancer 2020;123:1063–70. https://doi.org/10.1038/s41416-020-0978-4.

[2] Nwankwo O, Mekdash H, Sihono DSK, Wenz F, Glatting G. Knowledge-based radiation therapy (KBRT) treatment planning versus planning by experts: validation of a KBRT algorithm for prostate cancer treatment planning. Radiat Oncol 2015;10:111. https://doi.org/10.1186/s13014-015-0416-6.

[3] McIntosh C, Welch M, McNiven A, Jaffray DA, Purdie TG. Fully automated treatment planning for head and neck radiotherapy using a voxel-based dose prediction and dose mimicking method. Phys Med Biol 2017;62:5926–44. https://doi.org/10.1088/1361-6560/aa71f8.

[4] Schubert C, Waletzko O, Weiss C, Voelzke D, Toperim S, Roeser A, et al. Intercenter validation of a knowledge based model for automated planning of volumetric modulated arc therapy for prostate cancer. The experience of the German RapidPlan Consortium. PLoS One 2017;12:e0178034. https://doi.org/10.1371/journal.pone.0178034.

[5] Guillaume S, Charnomordic B. Generating an interpretable family of fuzzy partitions from data. IEEE Trans Fuzzy Syst 2004;12:324–35. https://doi.org/10.1109/TFUZZ.2004.825979.

[6] Guillaume S, Charnomordic B. Learning interpretable fuzzy inference systems with FisPro. Inf Sci 2011;181:4409–27. https://doi.org/10.1016/j.ins.2011.03.025.

[7] Chiu SL. Fuzzy model identification based on cluster estimation. J Intell Fuzzy Syst 1994;2:267–78. https://doi.org/10.3233/ifs-1994-2306.

[8] Marks LB, Yorke ED, Jackson A, Ten Haken RK, Constine LS, Eisbruch A, et al. Use of normal tissue complication probability models in the clinic. Int J Radiat Oncol Biol Phys 2010;76:S10–9. https://doi.org/10.1016/j.ijrobp.2009.07.1754.

[9] Dearnaley D, Syndikus I, Mossop H, Khoo V, Birtle A, Bloomfield D, et al. Conventional versus hypofractionated high-dose intensity-modulated radiotherapy for prostate cancer: 5-year outcomes of the randomised, non-inferiority, phase 3 CHHiP trial. Lancet Oncol 2016;17:1047–60. https://doi.org/10.1016/S1470-2045(16)30102-4.

[10] Arunsingh M, Mallick I, Prasath S, Arun B, Sarkar S, Shrimali RK, et al. Acute toxicity and its dosimetric correlates for high-risk prostate cancer treated with moderately hypofractionated radiotherapy. Med Dosim 2017;42:18–23. https://doi.org/10.1016/j.meddos.2016.10.002.

[11] Maulik S, Arunsingh M, Arun B, Prasath S, Mallick I. Moderately Hypofractionated Radiotherapy and Androgen Deprivation Therapy for High-risk Localised Prostate Cancer: Predictors of Long-term Biochemical Control and Toxicity. Clin Oncol 2021. https://doi.org/10.1016/j.clon.2021.08.010.

[12] Peduzzi P, Concato J, Feinstein AR, Holford TR. Importance of events per independent variable in proportional hazards regression analysis. II. Accuracy and precision of regression estimates. J Clin Epidemiol 1995;48:1503–10. https://doi.org/10.1016/0895-4356(95)00048-8.

[13] Zhang J, Wu QJ, Xie T, Sheng Y, Yin F-F, Ge Y. An Ensemble Approach to Knowledge-Based Intensity-Modulated Radiation Therapy Planning. Front Oncol 2018;8:57. https://doi.org/10.3389/fonc.2018.00057.

[14] Shoaip N, El-Sappagh S, Barakat S, Elmogy M. Chapter 7 - Ontology enhanced fuzzy clinical decision support system. In: Dey N, Ashour AS, Fong SJ, Borra S, editors. U-Healthcare Monitoring Systems, Academic Press; 2019, p. 147–77. https://doi.org/10.1016/B978-0-12-815370-3.00007-4.

[15] Kruskal WH, Wallis WA. Use of Ranks in One-Criterion Variance Analysis. J Am Stat Assoc 1952;47:583–621. https://doi.org/10.1080/01621459.1952.10483441.

[16] Hussein M, South CP, Barry MA, Adams EJ, Jordan TJ, Stewart AJ, et al. Clinical validation and benchmarking of knowledge-based IMRT and VMAT treatment planning in pelvic anatomy. Radiother Oncol 2016;120:473–9. https://doi.org/10.1016/j.radonc.2016.06.022.

[17] Ge Y, Wu QJ. Knowledge-based planning for intensity-modulated radiation therapy: A review of data-driven approaches. Med Phys 2019;46:2760–75. https://doi.org/10.1002/mp.13526.

[18] Wu B, Ricchetti F, Sanguineti G, Kazhdan M, Simari P, Jacques R, et al. Data-driven approach to generating achievable dose-volume histogram objectives in intensity-modulated radiotherapy planning. Int J Radiat Oncol Biol Phys 2011;79:1241–7. https://doi.org/10.1016/j.ijrobp.2010.05.026.

[19] Yuan L, Ge Y, Lee WR, Yin FF, Kirkpatrick JP, Wu QJ. Quantitative analysis of the factors which affect the interpatient organ-at-risk dose sparing variation in IMRT plans. Med Phys 2012;39:6868–78. https://doi.org/10.1118/1.4757927.

[20] Cagni E, Botti A, Micera R, Galeandro M, Sghedoni R, Orlandi M, et al. Knowledge-based treatment planning: An inter-technique and inter-system feasibility study for prostate cancer. Phys Med 2017;36:38–45. https://doi.org/10.1016/j.ejmp.2017.03.002.

[21] Ma M, Kovalchuk N, Buyyounouski MK, Xing L, Yang Y. Dosimetric features-driven machine learning model for DVH prediction in VMAT treatment planning. Med Phys 2019;46:857–67. https://doi.org/10.1002/mp.13334.

[22] Zhu X, Ge Y, Li T, Thongphiew D, Yin F-F, Wu QJ. A planning quality evaluation tool for prostate adaptive IMRT based on machine learning. Med Phys 2011;38:719–26. https://doi.org/10.1118/1.3539749.

[23] Appenzoller LM, Michalski JM, Thorstad WL, Mutic S, Moore KL. Predicting dose-volume histograms for organs-at-risk in IMRT planning. Med Phys 2012;39:7446–61. https://doi.org/10.1118/1.4761864.

[24] Fogliata A, Belosi F, Clivio A, Navarria P, Nicolini G, Scorsetti M, et al. On the pre-clinical validation of a commercial model-based optimisation engine: application to volumetric modulated arc therapy for patients with lung or prostate cancer. Radiother Oncol 2014;113:385–91. https://doi.org/10.1016/j.radonc.2014.11.009.

[25] Powis R, Bird A, Brennan M, Hinks S, Newman H, Reed K, et al. Clinical implementation of a knowledge based planning tool for prostate VMAT. Radiat Oncol 2017;12:81. https://doi.org/10.1186/s13014-017-0814-z.

[26] Mallick I, Saha S, Arunsingh MA. A Web-based Dose-volume Histogram Dashboard for Library-based Individualized Dose-constraints and Clinical Plan Evaluation. J Med Syst 2021;45:62. https://doi.org/10.1007/s10916-021-01740-9.

APPENDIX:

## **A. Development of fuzzy rule based prediction model:**

From the existing literature, it has been observed that the machine learning methods are inherently less interpretable. Additionally, models built by implementing regression techniques or machine learning techniques have not been found to produce results better than what they find in the training data [5]. This can lead to poor predictive performance for a new data set that is significantly different from the training set. When dealing with large delicate systems, like treatment planning problems, this is a serious drawback. Along with that, overfitting is another major issue faced with some machine learning algorithms. It occurs when a model gives more preference to handling random errors or noise instead of focusing on the underlying relationship of the features. All these flaws motivate us to implant a Fuzzy Rule-Based Prediction (FRBP) model in the dose-volume prediction problem. Rule-based learning algorithms provide a compact, flexible, and interpretable representation. It generalizes the problem statement from the presented data to model unseen situations in a reasonable way. Besides, it can lead to an acceptable result even for a limited data-set. Thus, the trade-off between accuracy and interpretability is the key to implementing the fuzzy rule-based prediction method.

To build an interpretable fuzzy rule base, fuzzy partitioning is a prerequisite condition. First, to form the fuzzy partition, we need to find the cluster center of the numerical data using a simple and effective algorithm based on a 'distance-like' function [6]. To search for the best fuzzy partition with the most appropriate number of clusters, we have used Hierarchical Fuzzy Partitioning (HFP) method [4,5] due to its advantages over the existing methods. In this implementation, we selected those good and steady partitions which have small standard deviations. For simplicity, we generated the fuzzy partition with semi trapezoidal shapes at the edges and triangular-shaped membership functions elsewhere. After splitting the input domain into regions, each fuzzy set is assigned a relevant linguistic level (e.g.: small, medium, high). This method could be carried out independently over any input dimension related to a feature variable.

After dividing each input domain into a fuzzy region using the HFP method, the next step is to form the fuzzy rule-base. Based on the assigned degrees of the features in different fuzzy sets, the primary fuzzy rule-base can be generated which contains all of the rules extracted from the training sample. Since in real-world problems with a huge amount of data, it is highly probable that there will be some conflicting rules, i.e., rules that have the same antecedent part with a different consequent label. Also, the rules extracted from the training data may include redundant structures as well as poorly performing rules. Thus, the construction of a compact fuzzy if-then rule-based for a given data is the next task. To enhance the overall performance of the model, the primary fuzzy rule-base needs to be refined by embracing medical experts' knowledge, who may decide to modify or delete some rules, or even add new ones.

Finally, to predict the DVH from fuzzy rule-based we have used the Fuzzy Decision Tree (FDT) model [5], which is an extension of the classical decision tree. The main advantage of implementing FDT is its comprehensibility and interpretability. This methodology uses a recursive procedure to split each node into requisite numbers of child nodes based on the gain function. It also ascertains a sorting order for influential variables, which can be useful for selecting the more relevant variables for further learning procedures.

## B. Analyzing the variance of errors in the models

|  | LR | RF | EN | SVR | MLP | FRBP |
|---|---|---|---|---|---|---|
| *Bladder (whole dose range)* | 1.962 | 6.617 | 3.863 | 6.421 | 2.019 | 4.501 |
| *Bladder (4000cGy-6420cGy)* | 0.055 | 3.099 | 0.200 | 1.747 | 0.054 | 1.128 |
| *Rectum (whole dose range)* | 3.805 | 2.635 | 3.497 | 3.632 | 3.090 | 3.325 |
| *Rectum (4000cGy-6420cGy)* | 0.955 | 1.229 | 0.927 | 1.148 | 1.047 | 1.662 |

*Tab.1: Summary of the Kruskal-Wallis test statistics value*

| Bladder | LR | RF | EN | SVR | MLP | FRBP |
|---|---|---|---|---|---|---|
| *3000cGy* | 0.051 | 0.019 | 0.020 | 0.002 | 0.051 | 0.007 |
| *4000cGy* | 0.700 | 0.377 | 0.671 | 0.996 | 0.708 | 0.248 |
| *4500cGy* | 0.753 | 0.045 | 0.621 | 0.163 | 0.708 | 0.275 |
| *5000cGy* | *0.944* | *0.067* | *0.784* | *0.128* | *0.936* | *0.332* |
| *5300cGy* | *0.936* | *0.057* | *0.814* | *0.125* | *0.992* | *0.378* |
| *5600cGy* | *0.980* | *0.068* | *0.686* | *0.143* | *0.897* | *0.302* |
| *5900cGy* | *0.846* | *0.028* | *0.439* | *0.189* | *0.924* | *0.218* |
| *6000cGy* | *0.433* | *0.031* | *0.123* | *0.296* | *0.610* | *0.406* |

*Tab.2: Summary of the statistical analysis (p-values) at particular dose range of bladder*

| *Rectum* | *LR* | *RF* | *EN* | *SVR* | *MLP* | *FRBP* |
|---|---|---|---|---|---|---|
| *3000cGy* | *0.118* | *0.371* | *0.176* | *0.463* | *0.128* | *0.280* |
| *4000cGy* | *0.065* | *0.061* | *0.056* | *0.087* | *0.076* | *0.008* |
| *4500cGy* | *0.909* | *0.265* | *0.768* | *0.642* | *0.846* | *0.077* |
| *5000cGy* | *0.964* | *0.532* | *0.948* | *0.539* | *0.916* | *0.524* |
| *5300cGy* | *0.715* | *0.678* | *0.853* | *0.893* | *0.657* | *0.798* |
| *5600cGy* | *0.135* | *0.210* | *0.203* | *0.085* | *0.118* | *0.219* |
| *5900cGy* | *0.433* | *0.885* | *0.822* | *0.096* | *0.366* | *0.136* |
| *6000cGy* | *0.138* | *0.182* | *0.145* | *0.822* | *0.350* | *0.246* |

*Tab.3: Summary of the statistical analysis (p-values) at particular dose range of rectum*

## C. Estimation of the confidence interval

As mentioned earlier, lower and upper bounds were determined using Weibull distribution. Before applying the Weibull distribution, first normality of the PTV (%) corresponding to each dose was checked by calculating skewness and kurtosis. Since the skewness was non-zero and kurtosis was other than three, the distribution was considered as non-Gaussian. Since PTVs (%) have been found to be non-Gaussian, the PTVs (%) have been fitted to the Weibull distribution. The probability density function (PDF) and cumulative distribution function (CDF) of the two-parameter Weibull distribution function are given by equation 1 and equation 2.

$$f(x) = \frac{k}{s}\left(\frac{x}{s}\right)^{k-1} exp\left[-\left(\frac{x}{s}\right)^{k}\right] \quad (1)$$

$$F(x) = 1 - exp\left[-\left(\frac{x}{s}\right)^{k}\right] \quad (2)$$

Where $x$ is PTV (%) corresponding to a particular dose, $k$ is the non-dimensional shape parameter and $s$ is the scale parameter. The dimensions of $s$ and $x$ are same. The least-square method (LSM) has been employed for estimating Weibull parameters (k and s). The double logarithm of the CDF (equation 2) yields equation 3.

$$ln\left(-ln(1-F(x))\right) = k\,ln(x) - k\,ln(s) \quad (3)$$

The plot of $ln\left(-ln(1-F(x))\right)$ against $ln(x)$ is used to estimate Weibull parameters by fitting a straight line $(y = ax + b)$ for the random variate $x$. The Weibull parameter $k$ is the slope of the best-fitted straight line and $s$ is given by the vertical intercept $(-k\,ln(s))$ i.e. $k = a$ and $s = exp(-b/a)$.

Further, we assumed a 95% confidence interval which gives lower and upper tails corresponding to each dose. The lower and upper bounds are determined by joining all lower tails and upper tails, respectively.

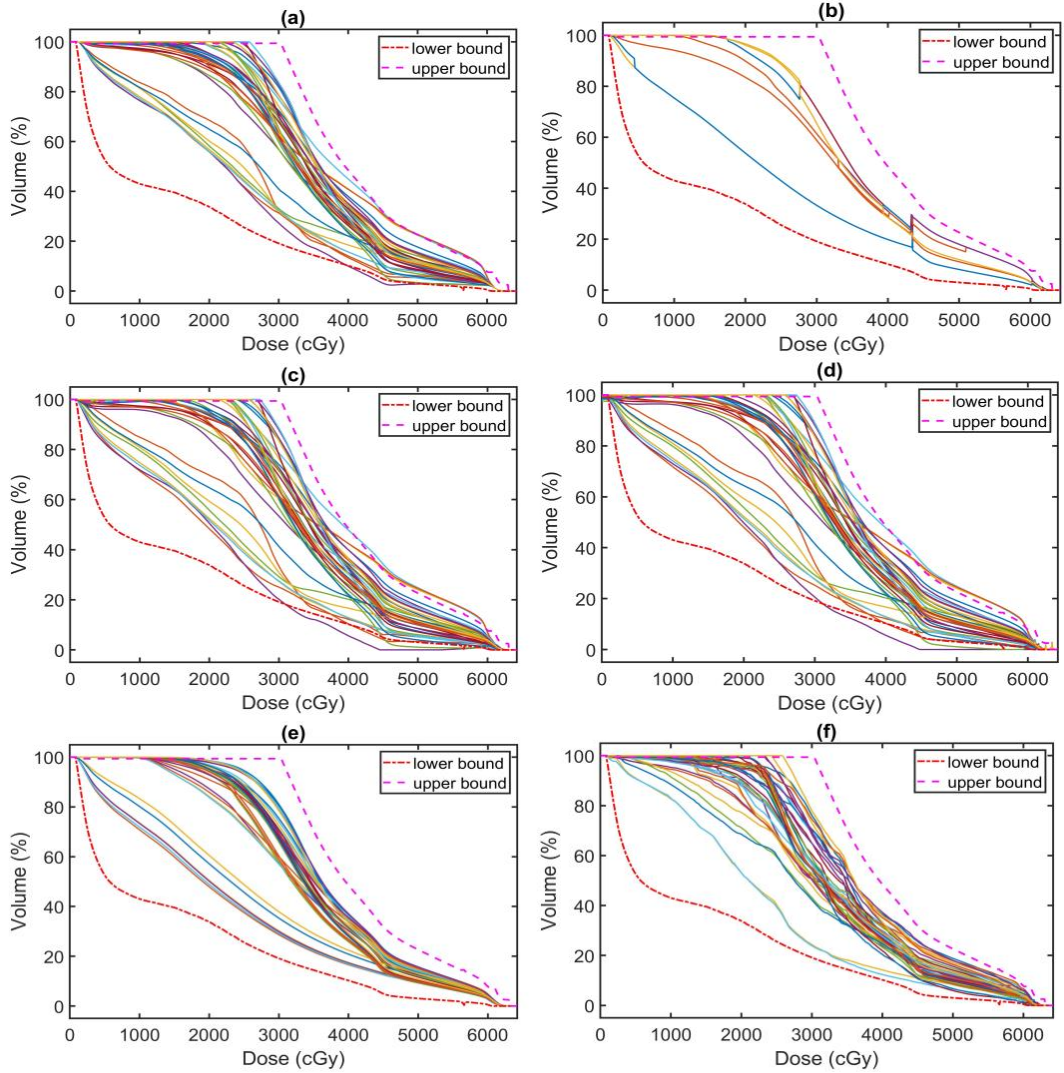

**Figure 4(a). Confidence interval for bladder for predicted DVH using different methods: (a) EN, (b) FRBP, (c) LR, (d) MLP, (e) RF, and (f) SVR**

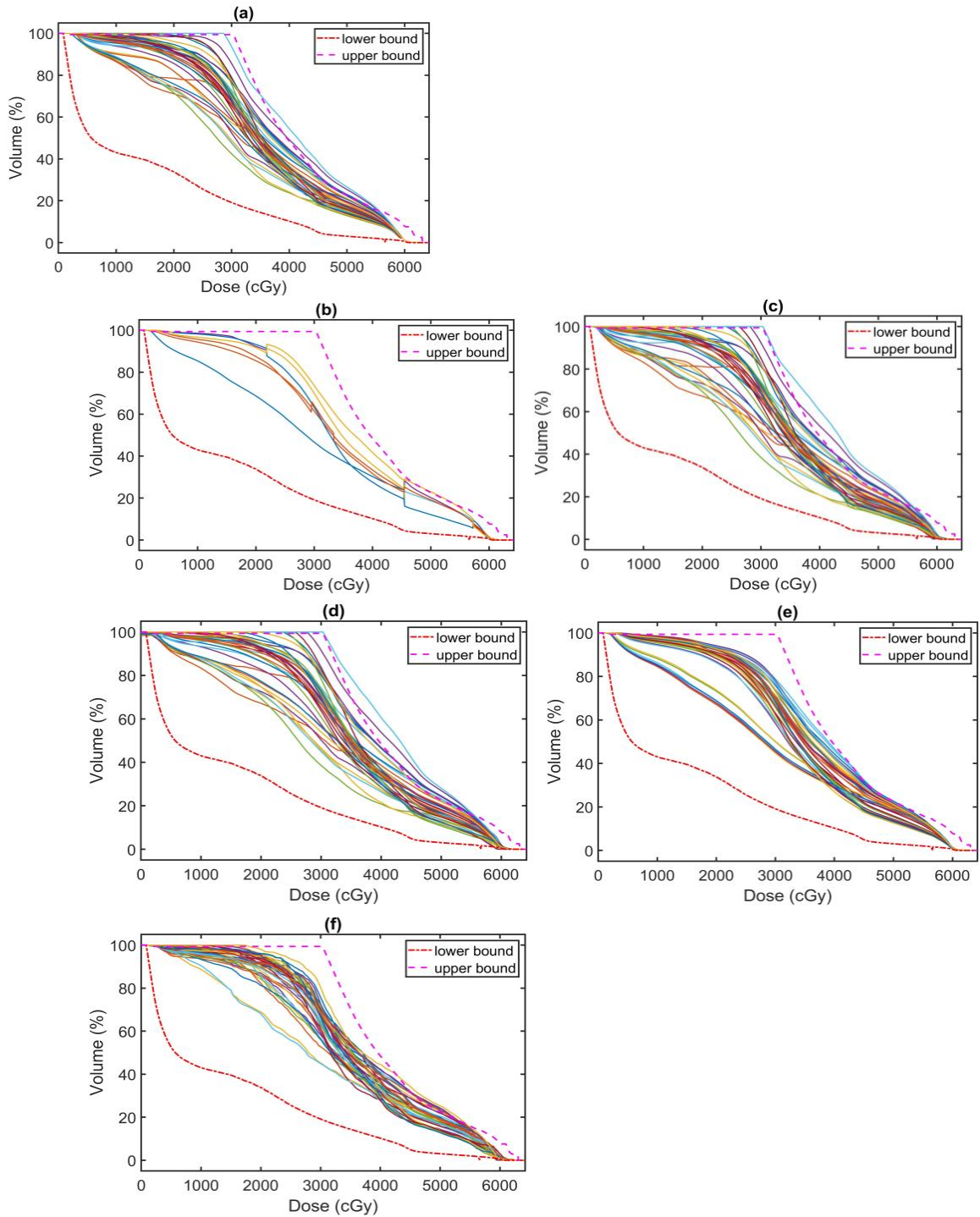

**Figure 4(b). Confidence interval for rectum for predicted DVH using different methods: (a) EN, (b) FRBP, (c) LR, (d) MLP, (e) RF, and (f) SVR**